\documentclass[10pt,twocolumn,letterpaper]{article}

\usepackage{iccv}
\usepackage[pagebackref=true,breaklinks=true,letterpaper=true,colorlinks,bookmarks=false]{hyperref}
\pdfoutput=1
\usepackage{times}
\usepackage{epsfig}
\usepackage{graphicx}
\usepackage{amsmath}
\usepackage{amssymb}
\usepackage{booktabs}
\usepackage{multirow}
\usepackage{bigstrut}
\usepackage{hyperref}
\usepackage{mathrsfs}
\usepackage{rotating}
\usepackage{adjustbox}
\usepackage[table]{xcolor}

\definecolor{my_green}{RGB}{18,159,87}

\usepackage{iccv}
\usepackage{times}
\usepackage{epsfig}
\usepackage{graphicx}
\usepackage{amsmath}
\usepackage{amssymb}
\usepackage{amsthm,amsmath,amssymb}

\usepackage{mathrsfs}

\usepackage[breaklinks=true,bookmarks=false]{hyperref}

\iccvfinalcopy 

\def\iccvPaperID{****} 

\ificcvfinal\fi

\begin{document}

\title{[CLS] Token is All You Need for Zero-Shot Semantic Segmentation}

\author{Letian Wu$^{1}$\footnotemark[1] \quad Wenyao Zhang$^{2,3}$ \quad  Tengping Jiang$^{4}$ \quad  Wankou Yang$^{1}$\footnotemark[2] \quad   Xin Jin$^{3}$\footnotemark[2]   \quad Wenjun Zeng$^{3}$ \\
$^{1}$Southeast University \quad
$^{2}$Shanghai Jiao Tong University \quad \\  $^{3}$Eastern Institute for Advanced Study \quad  
$^{4}$Wuhan University 
\\
}

\maketitle
\ificcvfinal\thispagestyle{empty}\fi
\renewcommand{\thefootnote}{\fnsymbol{footnote}}
\footnotetext[1]{This work was done when Letian Wu was an intern at Eastern Institute for Advanced Study.}
\footnotetext[2]{Corresponding author, Wankou Yang, \url{wkyang@seu.edu.cn}, Xin Jin, \url{jinxin@eias.ac.cn}}
\begin{abstract}
In this paper, we propose an embarrassingly simple yet highly effective zero-shot semantic segmentation (ZS3) method, based on the pre-trained vision-language model CLIP. First, our study provides a couple of key discoveries: (i) the global tokens (a.k.a [CLS] tokens in Transformer) of the text branch in CLIP provide a powerful representation of semantic information and (ii) these text-side [CLS] tokens can be regarded as category priors to guide CLIP visual encoder pay more attention on the corresponding region of interest. Based on that, we build upon the CLIP model as a backbone which we extend with a \textbf{One-Way [CLS] token navigation} from text to the visual branch that enables zero-shot dense prediction, dubbed \textbf{ClsCLIP}. Specifically, we use the [CLS] token output from the text branch, as an auxiliary semantic prompt, to replace the [CLS] token in shallow layers of the ViT-based visual encoder. This one-way navigation embeds such global category prior earlier and thus promotes semantic segmentation. Furthermore, to better segment tiny objects in ZS3, we further enhance ClsCLIP with a local zoom-in strategy, which employs a region proposal pre-processing and we get ClsCLIP+. Extensive experiments demonstrate that our proposed ZS3 method achieves a SOTA performance, and it is even comparable with those few-shot semantic segmentation methods.
\end{abstract}

\section{Introduction}
As a fundamental task in computer vision, semantic segmentation aims to predict the category of each pixel of an image. With the fast development of deep learning technology, fully-supervised semantic segmentation has achieved great success~\cite{FCN,deeplabv3,pspnet}. Despite unprecedented advances, existing methods have heavily relied on the availability of a large amount of annotated training images, which involves a substantial amount of labor. Besides, they often have poor generalization and usability in real-world deployments, and cannot perform well on unseen new categories due to the domain gaps. This gives rise to a surging interest in low-supervision-based semantic segmentation approaches, including semi-supervised~\cite{Zhong2021PixelCS}, weakly-supervised~\cite{Zhang2021ComplementaryPF}, few-shot~\cite{Min2021HypercorrelationSF}. However, the segmentation performance of these methods is still limited by the quality of those insufficient annotated samples. Thus, many efforts have resorted to exploring zero-shot semantic segmentation~\cite{bucher2019zero}.

Zero-shot semantic segmentation (ZS3) aims to segment novel unseen categories that have not been seen in the training without any extra annotations, which is extremely challenging due to the large gaps among different objects and less valuable information that can be leveraged. The core of ZS3 is twofold: (i) properly inheriting the well-learned knowledge of seen categories, and absorbing transferable information to complete ZS3; (ii) with the help of auxiliary semantic information (e.g. similar attribute features obtained by clustering or cross-modal tag word embeddings) to get the connections between seen and unseen categories and thus complete ZS3. Based on that, we analyze that a good solution to ZS3 should meet two basic requirements: (i) based on enough prior knowledge derived from seen data or open-world and (ii) has a human-level recognition ability w.r.t the semantic information.

Coincidentally, recent investigations on large-scale vision-language model (VLM) pre-training ~\cite{UNITER,clip} are exactly pursuing the above-mentioned goals by learning visual representation via textual supervision. Among them, the most representative work of CLIP~\cite{clip} encode image and texts separately and then map them into a unified space via a contrastive loss for pre-training. CLIP has demonstrated a powerful transferability on a variety of downstream tasks in
a zero-shot manner with some parameter-efficient transfer learning methods like prompt tuning~\cite{vpt,coop}. 
Thus, we wonder that can we leverage the rich generalized knowledge of CLIP, and navigate its visual encoder to perform zero-shot semantic segmentation with the help of an auxiliary semantic prompt provided by the other text branch.

\begin{figure*}[t]
	\centering
	\includegraphics[width=\linewidth]{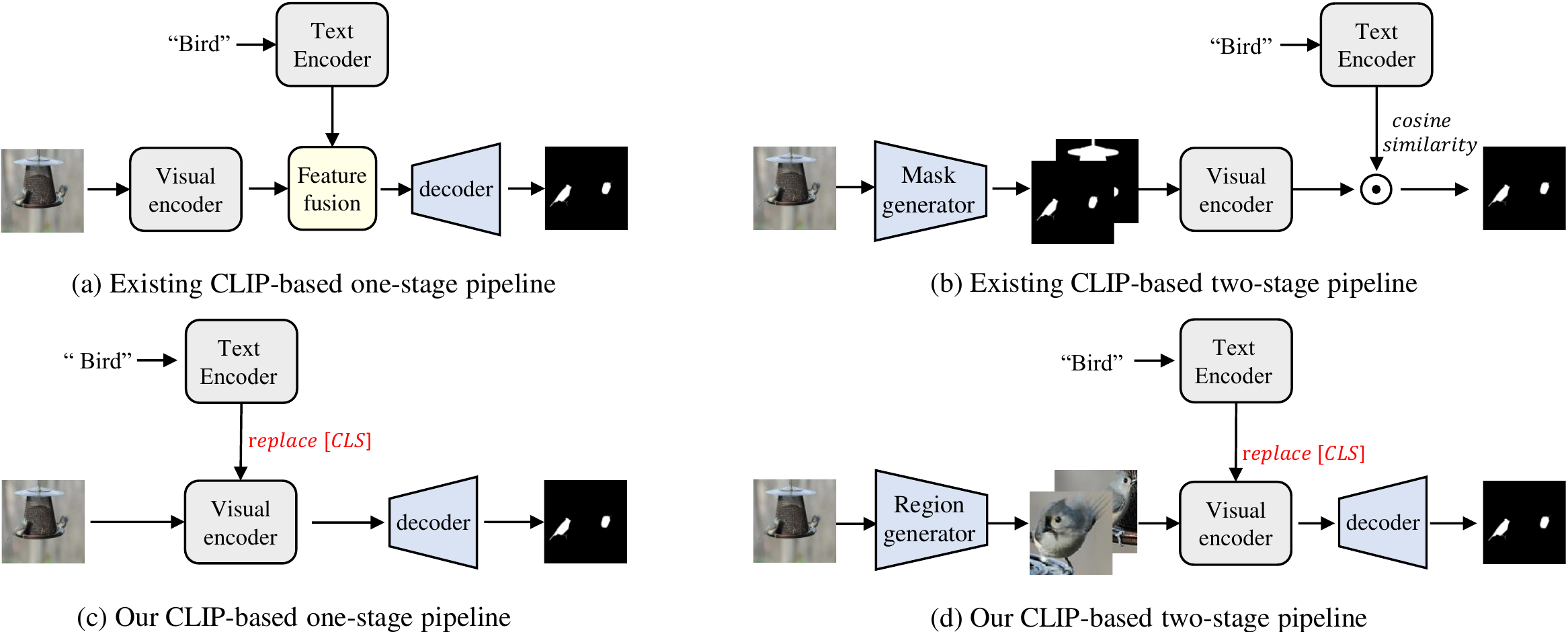}
	\caption{The difference between our proposed pipelines and other pipelines on zero-shot semantic segmentation. (a) The existing one-stage pipeline designs two modal fusion modules for feature maps to predict results. (b) The existing two-stage pipeline generates class-agnostic masks and then classifies them via CLIP. (c) Our proposed one-stage pipeline replace the visual-side [CLS] token with the text-side [CLS] token to guide the encoder to focus on the region of interest. (d) Our proposed two-stage pipeline provides the preliminary prior position information of the target to be segmented on the basis of the one-stage.}
	\label{fig1}
   \vspace{-0.4cm}
\end{figure*}

Obviously, the answer is ``Yes''. Some concurrent methods have been explored to apply CLIP into ZS3~\cite{Lseg,clipseg,zsseg,zegformer}, which can be grouped into two branches: one-stage CLIP-based ZS3 and two-stage CLIP-based ZS3. As shown in Figure.~\ref{fig1}(a), the existing schemes of one-stage CLIP-based ZS3~\cite{Lseg,clipseg,ZegCLIP} tend to roughly fuse two modal features together and send them to the subsequent decoder for dense prediction. Although this strategy leveraged the rich generalized knowledge of CLIP for ZS3, it did not explicitly and timely explore the potential of auxiliary semantic information derived from text-side. For the two-stage CLIP-based ZS3 methods~\cite{zsseg,zegformer} that are shown in Figure~\ref{fig1}(b), their pipelines tend to first generate masks with a segmentor and then classify these masks with CLIP~\cite{zsseg,zegformer}(Figure~\ref{fig1}.(b)), where CLIP is essentially taken as a classifier and has not yet fully unleashed its potential.

To alleviate the above shortcomings, in this paper, we extend the capabilities of CLIP from zero-shot classification to zero-shot semantic segmentation following the one-stage pipeline. In this way, CLIP actively takes part in the segmentation rather than just serving as a category classifier. Furthermore, we unleash the potential of CLIP by additionally utilizing the global semantic concept learned by the text branch for ZS3. Specifically, we build our ZS3 framework upon CLIP with a One-Way [CLS] token navigation, dubbed ClsCLIP. In detail, 
as a kind of auxiliary semantic prompt, we use the [CLS] tokens of the transformer-based encoder in the text branch to replace the [CLS] token in shallow layers of the ViT-based visual encoder. This one-way navigation design embeds such global category prior earlier into the visual encoder, which drives it to pay more attention to its interested regions. As a result, ClsCLIP naturally promotes zero-shot semantic segmentation. As shown in Figure~\ref{fig1}(c), ClsCLIP is the first work, to the best of our knowledge, to leverage the global semantic concept (i.e., text-side [CLS] tokens) learned by text branch for help ZS3 via replacing [CLS] tokens in CLIP ViT-based visual encoder. 
This operation is reasonable and its physical meaning lies on that--thanks to CLIP's unique image-text alignment and powerful generalization ability, the [CLS] token of text encoder has learned a strong category prior, so the one-way [CLS] token navigation embeds such global category prior into visual encoder earlier and thus promote the following dense prediction by the decoder. 

As a minor by-product, we further solve another common problem faced by ZS3 in this paper, that is, segmentor is prone to miss tiny objects. To this end, we further enhance ClsCLIP with a local zoom-in strategy and get ClsCLIP+, as shown in Figure~\ref{fig1}(d),  which employs a region proposal pre-processing, such as YOLO~\cite{YOLOv7}, first to affirm all the objects' locations that need to be segmented. Such pre-processing alleviates the issue of missing tiny objects segmentation results and also narrows down the following segmentation space. Note that, different from the previous two-stage ZS3 methods, ClsCLIP+ just uses a rough and lightweight pre-processing to get location proposals to make the entire pipeline more focused on interested segments, not like previous works directly output all objects' segmented masks and then classify them. 
All in all, we summarize our contributions of this paper as follows, 

\begin{itemize}
\item We extend the ability of CLIP zero-shot classification to zero-shot semantic segmentation with a one-way [CLS] token navigation strategy, and propose ClsCLIP, an embarrassingly simple yet highly effective ZS3 method. ClsCLIP replaces the [CLS] tokens in shallow layers of the ViT-based visual encoder with the text-side [CLS] tokens, which auxiliary semantic prompts help the subsequent dense prediction. 

\item We pinpoint that the one-way [CLS] token navigation in ClsCLIP is reasonable, and analyze that thanks to CLIP’s unique image-text alignment learning method, the [CLS] token of text encoder has learned
a strong category concept/prior. So, the one-way [CLS] token navigation embeds such global category prior earlier into the visual encoder, which indeed promotes the following dense prediction by decoder.

\item As a minor by-product contribution, we further employ a region proposal pre-processing to enhance ClsCLIP with object localization prior to being segmented. This scheme, called ClsCLIP+, alleviates the issue of missing tiny objects segmentation results.

\end{itemize} 

Extensive experiments on few-shot segmentation benchmarks with the zero-shot setting have demonstrated that our ClsCLIP and ClsCLIP+ both achieve state-of-the-art performance, and they are even superior to the few-shot semantic segmentation methods.

\section{Related Works}
\textbf{Pre-trained Vision Language Models.} 
Pre-trained Vision Language Models~\cite{UNITER,clip} connect image and text representations together. The most famous work is CLIP~\cite{clip}, which aligns the representation space of image and text via contrastive learning. Due to its powerful encoding capability, CLIP and its variants perform well on many downstream tasks, such as sketch synthesis~\cite{CLIPassoSO}, dense prediction~\cite{DenseCLIP}, zero-shot semantic segmentation~\cite{zsseg,zegformer}, reference segmentation~\cite{ReSTRCR}, object detection~\cite{RegionCLIPRL}. 
For zero-shot semantic segmentation, the current two-stage pipeline essentially uses the zero-shot classification ability of CLIP~\cite{zsseg,zegformer}.
~\cite{ZegCLIP} suggested a one-stage CLIP-based method for ZS3, but the modules are overly meticulously designed. 
Based on prompt learning, we propose a one-way [CLS] token navigation to transfer the ability of CLIP zero-shot classification to zero-shot semantic segmentation. The method that comes closest to our work is CLIPSeg~\cite{clipseg}, which modulates the decoder by FiLM~\cite{Film} using the output embedding of the text branch as a conditional vector. The idea diverges significantly from ours. We use the text-side [CLS] token instead of the embedding vector as an auxiliary semantic prompt to guide the encoder rather than through the conditional vector modem decoder. \par
\textbf{Generalized semantic segmentation.} The semantic segmentation task is to make pixel-by-pixel category predictions for an image on a fixed category label~\cite{FCN}. Few-shot semantic segmentation is to relax the restrictions of fixed categories~\cite{PASCAL5i}. By providing several images of new categories and pixel-level annotations (support set), the dense prediction ability of the network can be generalized to new categories. However, few-shot semantic segmentation needs to provide at least one annotation of new categories. 
The purpose of zero-shot semantic segmentation is to enable the model to predict new categories exclusively from auxiliary semantic information without any extra pixel-level annotations of new categories~\cite{bucher2019zero}, further relaxing the annotation requirements for new categories. 
Therefore, the model must establish the correlation between the image and auxiliary semantic information from seen categories. 
The rise of CLIP provides a fresh perspective on the interaction between auxiliary semantic information and image information.\par
\textbf{Prompt Learning.} Prompt learning was first widely used in NLP~\cite{gpt}. The initial approach was to design different hard prompts to bridge the gap between the downstream task and the pre-trained model, guiding the model to complete the corresponding downstream task~\cite{gpt,gpt2,gpt3}. 
Soon, some learnable parameters or modules are introduced as soft prompts to improve the transfer effect~\cite{prompttuning,vpt}. Our proposed method uses the CLIP text-side [CLS] tokens as the auxiliary semantic prompt to drive the ViT-based visual encoder to pay more attention to its interested regions.\par
\textbf{Instance Segmentation.} Instance segmentation tasks aim to distinguish different instances of certain classes in an image at the pixel level. In the candidate-box-based instance segmentation method, the candidate boxes are used to identify various instances, and then the mask segmentation is performed on the found object~\cite{Deepmask,InstanceSensitiveFCN,MaskRCNN}. Our proposed method has certain similarities with the idea of strength segmentation, but instead of obtaining various class instances, our aim is to use the region proposal generator as a pre-processing to first give preliminary prior position information about the object.\par
\section{Method}

\subsection{Problem Definition}
The goal of ZS3 is to learn from seen classes $\mathcal{C}_s$ and their pixel-level masks $\mathbf{M}_s$, and to predict masks for unseen classes $\mathcal{C}_u$. Note that seen and unseen classes do not intersect, i.e. $\mathcal{C}_s\cap \mathcal{C}_u=\varnothing$.
In general, the idea of ZS3 is to train the model to understand the mapping between the two modalities by using image feature samples in the seen classes and corresponding auxiliary semantic words (such as \textit{dog}, \textit{table}).
Then the model can transfer the learned knowledge to the unseen classes by using their auxiliary semantic information for dense prediction.\par


There are several settings for ZS3~\cite{bucher2019zero,Gu2020ContextawareFG} based on the various protocols for zero-shot semantic segmentation.
And we adopt the ``inductive'' setting which indicates the image of the novel semantic class and corresponding auxiliary semantic word are not available at the training stage. 
Formulaically, the training set $\mathbf{X}_{train}=\{(\mathbf{I}_{k},\mathbf{M}_{k},\mathbf{A}_{k})\}$ consists of input images $\mathbf{I}_{k}$, pixel-level masks $\mathbf{M}_{k}$ and auxiliary semantic words $\mathbf{A}_{k}$.
The test set $\mathbf{X}_{test}$ is unavailable in the training stage.
Note that the masks $\mathbf{M}_{k}$ of the training set $\mathbf{X}_{train}$ only cover the seen classes, while the unseen classes have to be predicted in the test set $\mathbf{X}_{test}$.
\begin{figure}[t]
	\centering
	\includegraphics[width=\linewidth]{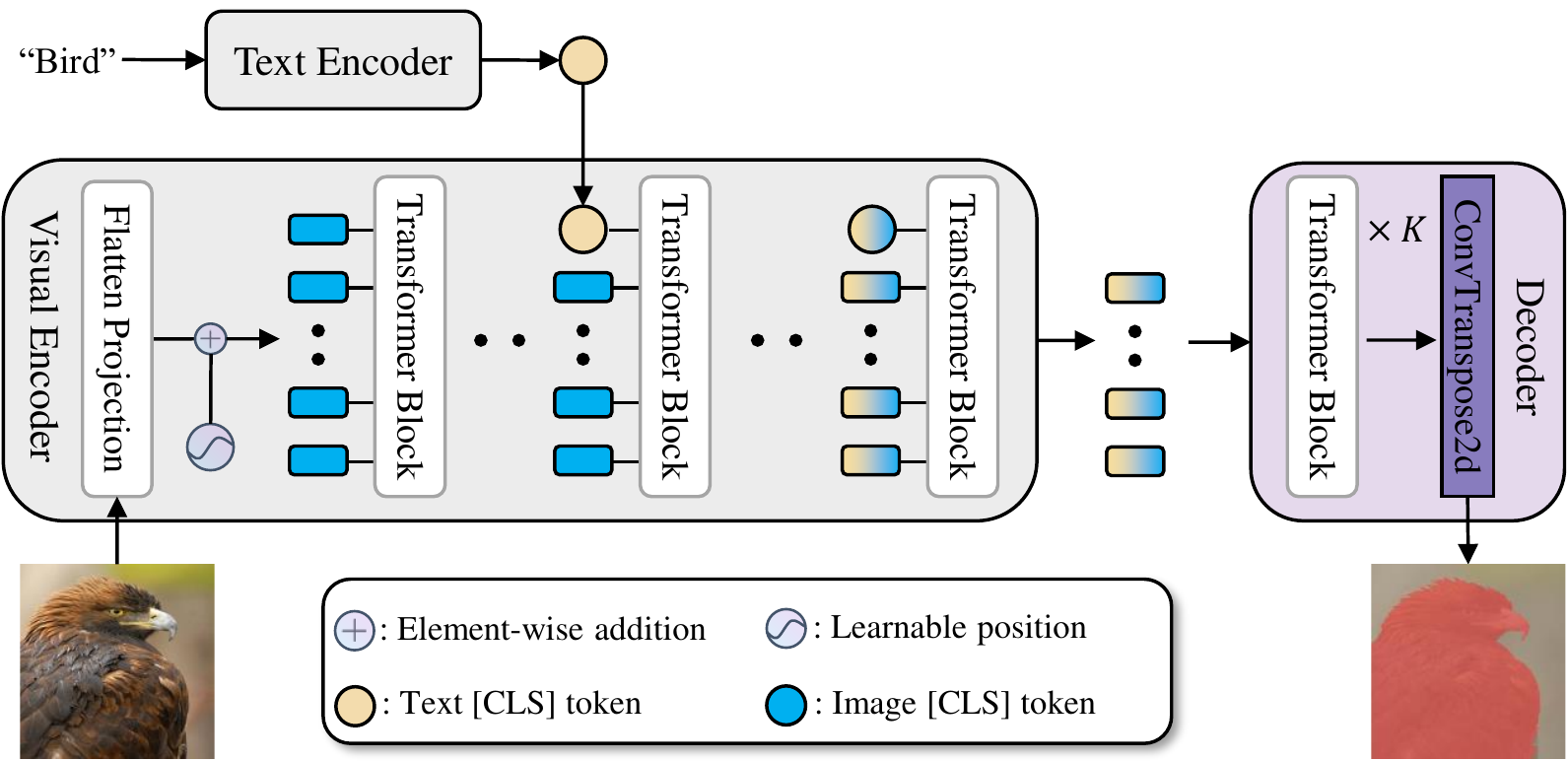}
	\caption{
 Overview of our proposed ClsCLIP method. ClsCLIP replaces the [CLS] token in the shallow layers of the visual encoder with the [CLS] token obtained from the text with strong semantic information. Then through the subsequent self-attention mechanism, the interaction and fusion between the two modalities are realized. 
 }
	\label{fig2}
  \vspace{-0.2cm}
\end{figure}

\subsection{A Revisit of CLIP}
\label{section:Revisiting}
A vanilla CLIP consists of a visual encoder, which is a ViT~\cite{vit} with $N$ layers, and a text encoder, which is a standard Transformer~\cite{transformer}.
Let $\mathbf{A}$ denote the auxiliary semantic word that serves as input to the CLIP text encoder, which will output a category [CLS] token $\mathbf{T}^{cls}$ according to Eq. \ref{equ1}.
\begin{equation}\label{equ1}
	\mathbf{T}^{cls}=\mathtt{F}_{t}(\mathbf{A}) \quad \mathbf{T}_{cls} \in \mathbb{R}^d,
\end{equation}
where $\mathtt{F}_{t}$ means the text encoder.
In zero-shot classification, the token embedding $\mathbf{T}^{cls}$ is usually projected further to compute its cosine similarity with the image embedding.\par
For the CLIP ViT-based visual encoder with $N$ layers, an input image $\mathbf{I}$ is first divided into $m$ fixed-sized patches $\{\mathbf{I}_j \in \mathbb{R}^{3\times h\times w} \mid j \in \mathbb{N},1\leq j\leq m\}$, where $h$ and $w$ are the image patch size. 
Then each patch is embedded into a $d$-dimensional latent feature space with positional encoding as the input to the visual encoder.
\begin{equation}\label{equ2}
        \mathbf{E}^{j}_{0}= \mathtt{Embed}(\mathbf{I}_j) \quad \mathbf{E}^{j}_{0} \in \mathbb{R}^d,j=1,2,3,...m \quad
\end{equation}\par
We denote by the set $\mathbf{E}_i$ the input to the $i^{th}$ layer of the visual encoder, i.e. $\mathbf{E}_i=\{\mathbf{E}^{j}_{i} \mid  1\leq j\leq m \}$. An additional learnable classification token ([CLS] token) $\mathbf{I}^{cls}_{0}$ is concatenate with $\mathbf{E}_0$ before the layer 0 input. 
The visual encoder output an embedding vector $\mathbf{y}$ based on Eq.~\ref{equ3} and Eq.~\ref{equ4}, 
\begin{align}
\label{equ3}
        [\mathbf{I}^{cls}_{i+1},\mathbf{E}_{i+1}] &=\mathtt{Layer}^{i}([\mathbf{I}^{cls}_{i},\mathbf{E}_i]) \quad 0\leq i\textless N-1\\
\label{equ4}
        \mathbf{y} &=\mathtt{Head}(\mathbf{I}^{cls}_{N}),
\end{align}
where $\mathbf{I}^{cls}_{i} \in \mathbb{R}^d $ and $\mathtt{Layer}^{i}$ means the $i^{th}$ layer of the visual encoder.\par
The self-attention mechanism in ViT enables information flow among different patches, allowing the [CLS] token to learn the image information about the category of the image. Therefore, the output embedding vector $\mathbf{y}$ with rich image information can be used to compute the cosine similarity with text embedding for subsequent classification.

\subsection{Proposed ZS3 Framework: ClsCLIP}
Derived from~\cite{clstoken}, an object can be reconstructed from high semantic information of global tokens (a.k.a [CLS] tokens in ViT). 
Since CLIP is trained by 400M image-text pairs via contrastive learning~\cite{clip}, the global tokens (a.k.a [CLS] tokens in Transformer) of the text branch in CLIP provide a powerful representation of semantic information, and the text-side [CLS] token and visual-side VIT-based [CLS] token are aligned on the latent space. 
Inspired by the above two points, we can conclude that these text-side [CLS] tokens can be regarded as a category prior to guide CLIP visual encoder to pay more attention to the corresponding region of interest.

\begin{figure}[t]
	\centering
	\includegraphics[width=\linewidth]{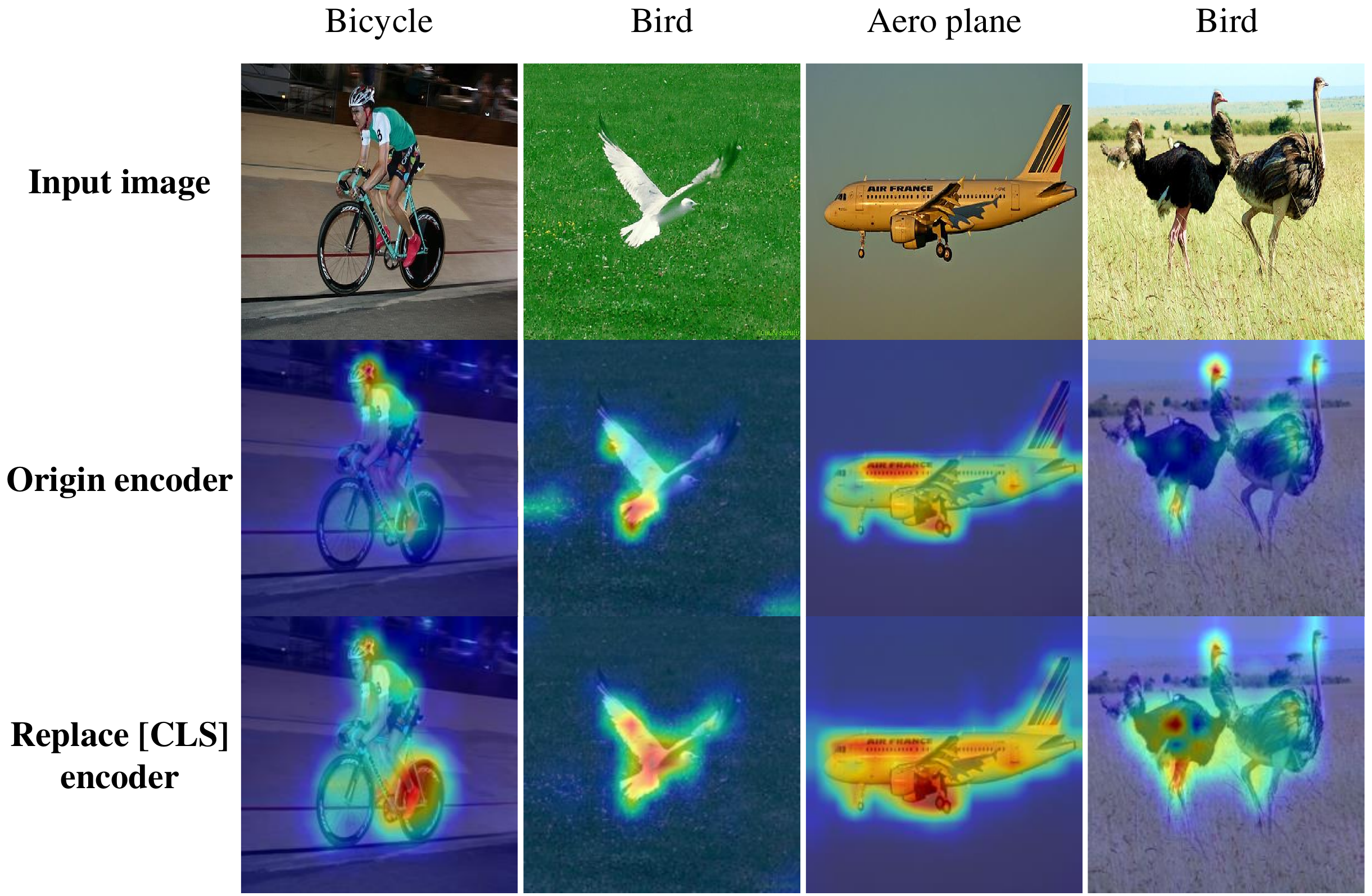}
	\caption{Comparison of the attention map of the visual encoder replacing the corresponding category [CLS] token with the original visual encoder. For example, when replaced by the ``bicycle'' text-side [CLS] token, the visual encoder focuses more on the bicycle-related area.}
	\label{fig:attention}
 \vspace{-0.2cm}
\end{figure}

The interaction of the self-attention mechanism in ViT can promote the information blending between image patches, so the [CLS] token $\mathbf{I}^{cls}_i$ can learn the information of the image, as detailed in Section~\ref{section:Revisiting}.
In other words, combining the mentioned conclusion, if the visual-side [CLS] token $\mathbf{I}^{cls}_i$ is replaced by $\mathbf{T}^{cls}$ with strong category information, the image patches may pay more attention to the region related to the category.\par
\begin{figure}[t]
	\centering
	\includegraphics[width=\linewidth]{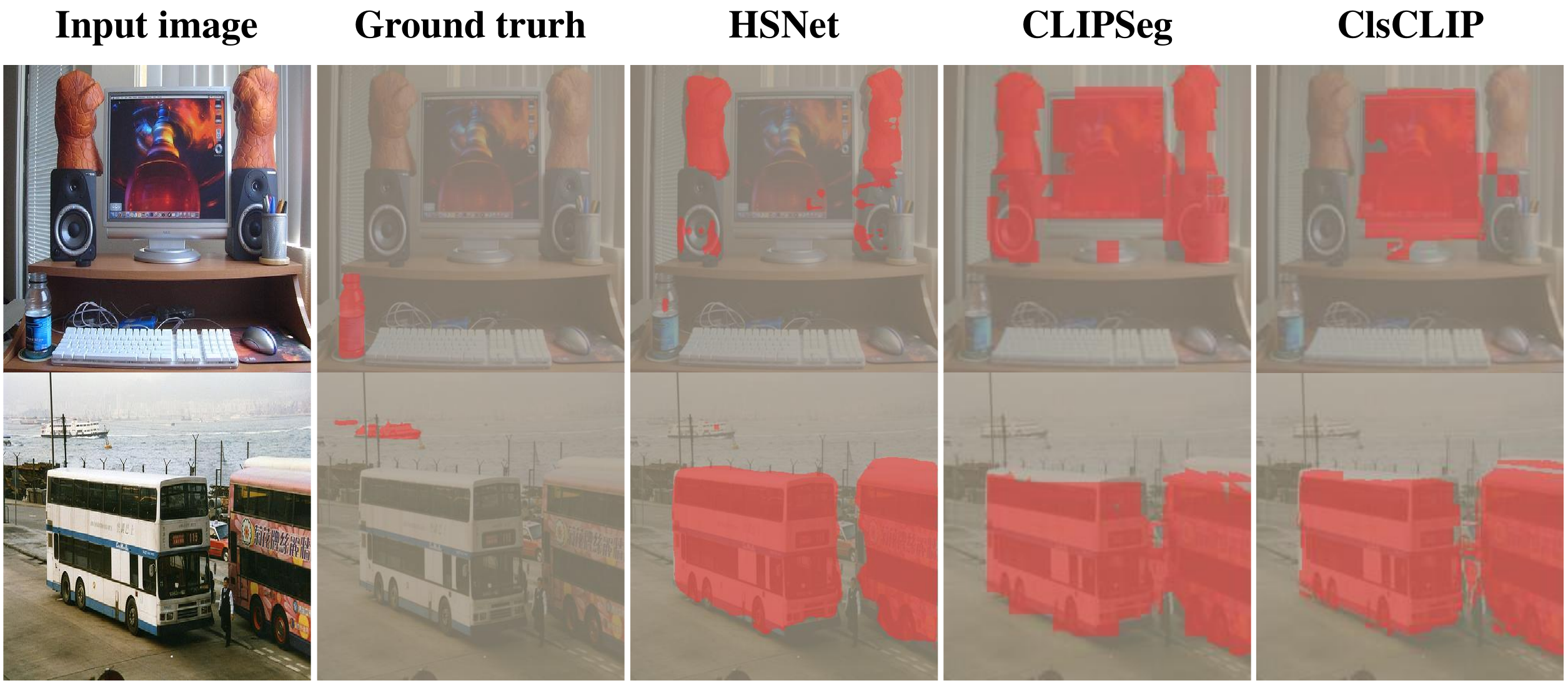}
	\caption{Examples of missing segmentation of tiny objects (bottle and boat) by HSNet (one-shot), CLIPSeg (zero-shot) and ClsCLIP (zero-shot).}
	\label{fig3}
  \vspace{-0.4cm}
\end{figure}
We conduct a simple experiment where we replace the shallow [CLS] tokens of the ViT-based visual encoder with the corresponding category text-side [CLS] tokens, and visualize the attention maps of the initial visual encoder and the modified visual encoder to reflect the attention regions of the image. 
Figure~\ref{fig:attention} shows that the modified visual encoder can focus more on the category-related information than the original visual encoder, which supports our hypothesis.
Therefore, we propose a one-way [CLS] token navigation to extend the processing power of CLIP image-level to pixel-level, which is embarrassingly simple yet highly effective, dubbed ClsCLIP, and Figure \ref{fig2} depicts its overall structure.
Formally, the above process is shown by Eq.~\ref{equ5} -- Eq.~\ref{equ7}.
\begin{align}
\label{equ5}
        [\mathbf{I}^{cls}_{i+1},\mathbf{E}_{i+1}] &= \mathtt{Layer}^{i}([\mathbf{I}^{cls}_{i},\mathbf{E}_{i}]) && 0\leq i\textless N_1 \\
\label{equ6}
 [\mathbf{I}^{cls}_{i+1},\mathbf{E}_{i+1}] &= \mathtt{Layer}^{i}([\mathtt{L}^{i}(\mathbf{T}^{cls}),\mathbf{E}_{i}]) && N_1\leq i\textless N_2 \\
\label{equ7}
         [\mathbf{I}^{cls}_{i+1},\mathbf{E}_{i+1}] &= \mathtt{Layer}^{i}([\mathbf{I}^{cls}_{i},\mathbf{E}_{i}]) && N_2\leq i\textless N,
\end{align}
where $\mathtt{L}_{i}$ means linear projection.\par

The output $\mathbf{E}_N=\{\mathbf{E}^{j}_{N} | 1\leq j\leq m \}$ of the last layer contains rich spatial information. 
Since $\mathbf{T}^{cls}$ has category priors, it prompts $\mathbf{E}_N$ to pay more attention to the region related to the category. 
We then feed $\mathbf{E}_N$ into a lightweight decoder composed of K-layer transformer to obtain the final segmentation result, as shown in Eq.~\ref{equ8}.

\begin{equation}\label{equ8}
        \mathbf{O}=\mathtt{Decoder}(\mathbf{E}_{N})
\end{equation}

\subsection{Advanced Version of ClsCLIP: ClsCLIP+}

We find a common problem faced in zero-shot and few-shot segmentation, that is, the segmentor is prone to miss tiny objects. 
Existing methods tend to prefer to segment other non-required large objects in the image more than tiny objects, and ClsCLIP suffers from the same problem, as shown in Figure~\ref{fig3}. 
The potential reason is that the low pixel ratio of tiny objects makes $\mathbf{T}^{cls}$ an inadequate semantic prompt which is challenging for $\mathbf{T}^{cls}$ to guide the visual encoder to focus on the object of interest .\par

\begin{figure}[t]
	\centering
	\includegraphics[width=\linewidth]{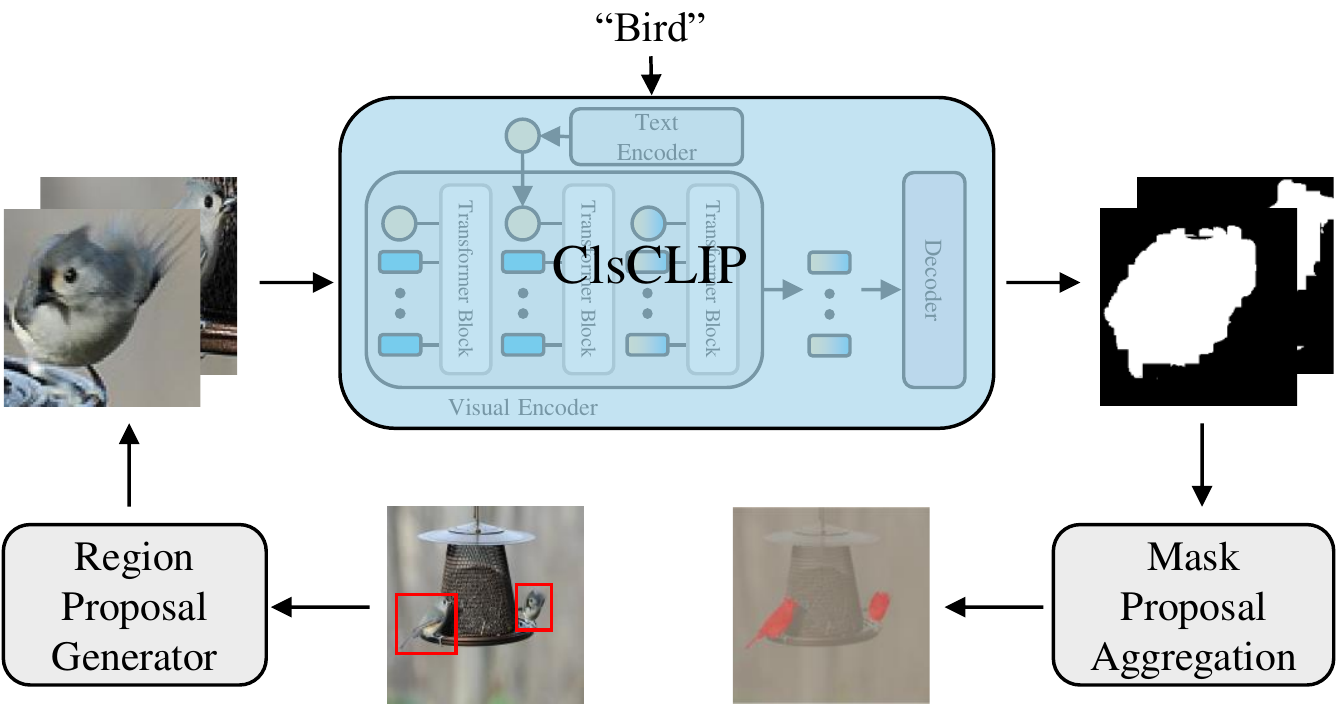}
	\caption{General architecture diagram of ClsCLIP+. Given an input image, we first use a region proposal generator to produce a set of candidate regions that may contain objects of interest. Then, for each region proposal, we apply ClSCLIP to generate a cluster of mask proposals that represent different semantic concepts. Finally, we aggregate the mask proposals across all regions to obtain the final pixel-level output.}
	\label{fig:clsclip+_v4}
   \vspace{-0.4cm}
\end{figure}

Therefore, we propose a local zoom-in strategy that employs a region proposal generator to locate all the objects of interest in the image, and then feed the extracted regions to ClsCLIP. 
This method, which we call ClsCLIP+, provides ClsCLIP with prior information about the object locations and reduces the segmentation search space. 
The overall framework of ClsCLIP+ is illustrated in Figure~\ref{fig:clsclip+_v4}.

Unlike the previous two-stage ZS3 methods that first produce segmentation masks for all objects and then classify them, ClsCLIP+ only uses a simple and lightweight pre-processing to obtain location proposals, which allows the entire framework to focus more on the segments of interest.\par
 
Concretely, we employ YOLOv7~\cite{YOLOv7} as a region proposal generator for the input image $\mathbf{I}$, producing a set of region proposals $\mathbf{P}_k=\{(\mathbf{R}_i,\mathbf{C}_i)\}$, where $\mathbf{R}_i$ and $\mathbf{C}_i$ represent the region proposal and its corresponding category, respectively.
Subsequently, we filter the region proposals using a category auxiliary vocabulary $\mathbf{A}_k$ that specifies the desired segmentation, resulting in a set of regions to be segmented denoted by $\mathbf{Q}_k=\{\mathbf{R}_m\mid \mathbf{C}_m=\mathbf{A}_k\}$.
Eventually, we apply ClsCLIP to segment each region in $\mathbf{Q}_k$, yielding a set of segmentation results denoted by $\mathbf{O}_k=\{\mathbf{O}_m\}$.
However, due to potential overlap between different regions, some pixels in the final output may have different segmentation results in different regions. Therefore, we adopt the aggregation principle of the "or operation" to determine the final result.
Specifically,  the class of a given pixel $q$ is determined by the following equation:
\begin{equation}\label{equ9}
        \mathbf{O}(q) = \bigcup_{i=1}^{n} \mathbf{O}_m(q)
\end{equation}\par

\section{Exploration Study of ClsCLIP}
In this section, we conduct extensive experiments on the PASCAL-$5^i$~\cite{PASCAL5i} to demonstrate the feasibility of our one-way [CLS] token navigation by answering a number of questions. 
The introduction and evaluation protocol of the dataset are detailed in Section \ref{section:Experiments}.\par

\textbf{Which way of using the text-side [CLS] token is better for ClsCLIP?} 
In our one-way [CLS] token navigation, we use text [CLS] token as an auxiliary prompt to replace the [CLS] token in some shallow layers of the ViT-based visual encoder to guide the image segmentation. 
Another common way to use text-side [CLS] token is to use it as attention weights to weight feature maps output by the visual encoder, including channel attention~\cite{channel} and spatial attention~\cite{spatial}.
As shown in Table \ref{table1},  these approaches do not perform well because they introduce the guidance information too late in the model,  which limits the dense prediction ability. Our method, on the other hand, effectively guides the model to focus on the region of interest from early stages.\par

\begin{table}[t]
  \centering
   \small
  \caption{Evaluation results for different ways of using the text [CLS] token.}
    \begin{tabular}{ccc}
    \toprule
 \makebox[0.12\textwidth][c]{Ways} & \makebox[0.09\textwidth][c]{mIoU (\%)} & \makebox[0.09\textwidth][c]{FB-IOU (\%)}\tabularnewline
    \midrule
 Channel attention~\cite{channel} & 68.3  & 78.6\tabularnewline
           Spatial attention~\cite{spatial} & 60.5  & 73.8 \tabularnewline
           \textbf{Replace [CLS] token (Ours)}  & \textbf{71.5} & \textbf{80.6} \tabularnewline
    \bottomrule
    \end{tabular}%
  \label{table1}%
\end{table}%

\begin{table}[t]
  \centering
   \small
  \caption{Evaluation results for different encoder update methods.}
    \begin{tabular}{ccc}
    \toprule
 \makebox[0.12\textwidth][c]{Encoder update } & \makebox[0.09\textwidth][c]{mIoU (\%)} & \makebox[0.09\textwidth][c]{FB-IOU (\%)}\tabularnewline
    \midrule
               VPT~\cite{vpt} & 64.7  & 76.3 \tabularnewline
   \textbf{Replace [CLS] token (Ours)}  & \textbf{71.5} & \textbf{80.6} \tabularnewline

    \bottomrule
    \end{tabular}%
    \vspace{-3.0mm}
  \label{table2}%
\end{table}%

\textbf{Which encoder update method is better for ClsCLIP? }
We freeze the weights of the visual encoder and use the [CLS] token as an auxiliary semantic prompt to guide the encoder to focus on the region of interest. 
A common alternative method is visual prompt tuning (VPT)~\cite{vpt}, which inserts learnable patches into the encoder and has been used~\cite{ZegCLIP} for ZS3. In our experiments, we set the number of learnable patches for VPT as 50. 
Table \ref{table2} shows that VPT reduces the dense prediction ability for unseen categories, which validates the effectiveness of our one-way [CLS] token navigation.

\textbf{How layers of the visual encoder should be replaced by the text [CLS] token as the prompt?}
We use CLIP-ViT-B/16 as backbone with a total of 12 layers for ClsCLIP and experiment with four different combinations of layers for replacing the text [CLS] token, as shown in Table \ref{table3}. 
The results show that replacing the original [CLS] token at the beginning is ineffective.
This may be due to the frozen encoder parameters, which prevent proper semantic alignment when the visual encoder’s original [CLS] token is simply replaced. 
Conversely, replacing the [CLS] token in the last three layers is also suboptimal because it introduces semantic information too late. 
Therefore, we conclude that it is preferable to replace the prompt at the shallow layers but not at the beginning.
This strategy avoids both semantic misalignment caused by early replacement and information loss caused by late replacement.

\begin{table}[t]
  \centering
   \small
  \caption{Evaluation results for different layers to replace the text [CLS] token.}
    \begin{tabular}{ccc}
\toprule
 \makebox[0.13\textwidth][c]{Layers} & \makebox[0.13\textwidth][c]{mIoU (\%)} & \makebox[0.14\textwidth][c]{FB-IOU (\%)}\tabularnewline
\midrule
 0,1,2 & 66.9  & 77.9 \tabularnewline
           2,5,8 & 67.0  & 79.0 \tabularnewline
           9,10,11 & 59.8  & 74.4 \tabularnewline
           \textbf{2,3,4 (Ours)}  & \textbf{71.5} & \textbf{80.6} \tabularnewline
    \bottomrule
    \end{tabular}%
        \vspace{-3.0mm}
  \label{table3}%
\end{table}%

\begin{table*}[t]
    \centering
    \small
    \caption{Evaluation results of mIoU and FB-IoU on PASCAL-$5^i$, \textcolor[rgb]{ 1,  0,  0}{\textbf{red}} and \textcolor[rgb]{ 0,  0,  1}{\textbf{blue}} indicate the best two scores. \textbf{Note that both ClsCLIP and ClsCLIP+ are zero-shot semantic segmantation methods.} ClsCLIP+ \textbf{with zero-shot setting} is even better than other 1-shot segmentation methods.}
           
    {(a). \textbf{Zero-shot method} ClsCLIP vs. Other zero-shot methods}{
       \vspace{0.1cm}
        \begin{tabular}{clccccccc}
        \toprule
        \makebox[0.03\textwidth][c]{} &
            \makebox[0.08\textwidth][l]{Method} & \makebox[0.1\textwidth][c]{Visual Backbone} & \makebox[0.1\textwidth][c]{Fold-0} & \makebox[0.09\textwidth][c]{Fold-1} & \makebox[0.09\textwidth][c]{Fold-2} & \makebox[0.09\textwidth][c]{Fold-3} & \makebox[0.09\textwidth][c]{Mean} & \makebox[0.09\textwidth][c]{FB-IOU}\tabularnewline
                    \midrule
   \multirow{5}[4]{*}{\begin{sideways}zero-shot\end{sideways}} & SPNet~\cite{SPNet} & \multirow{2}{*}{ResNet101}  & 23.8  & 17.0    & 14.1  & 18.3  & 18.3  & 44.3 \tabularnewline
   & ZS3Net~\cite{bucher2019zero} &              & 40.8  & 39.4  & 39.3  & 33.6  & 38.3  & 57.7 \tabularnewline
    \cmidrule{2-9}

   & LSeg~\cite{Lseg} & ViT-L/16  & \textcolor[rgb]{ 1,  0,  0}{\textbf{61.3}}  & \textcolor[rgb]{ 1,  0,  0}{\textbf{63.6}}  & \textcolor[rgb]{ 0,  0,  1}{\textbf{43.1}}  & 41.0    & \textcolor[rgb]{ 0,  0,  1}{\textbf{52.3}}  & \textcolor[rgb]{ 0,  0,  1}{\textbf{67.0}} \tabularnewline
  &  CLIPSeg~\cite{clipseg} & ViT-B/16  & 53.9 & \textcolor[rgb]{ 0,  0,  1}{\textbf{62.0}} & 42.8 & \textcolor[rgb]{ 0,  0,  1}{\textbf{48.0}} & 51.7  & 66.2 \tabularnewline
    \cmidrule{2-9}
  &  \textbf{ClsCLIP (Ours)} & ViT-B/16  & \textcolor[rgb]{ 0,  0,  1}{\textbf{59.9}} & 61.2 & \textcolor[rgb]{ 1,  0,  0}{\textbf{50.5}} & \textcolor[rgb]{ 1,  0,  0}{\textbf{54.1}} & \textcolor[rgb]{ 1,  0,  0}{\textbf{56.4}}  & \textcolor[rgb]{ 1,  0,  0}{\textbf{67.9}} \tabularnewline

\bottomrule
    \end{tabular}
    
    } 
    \vspace{0.2cm}
        {(b). \textbf{Zero-shot method} ClsCLIP+ vs. Other 1-shot 
        methods.}{
      
    \begin{tabular}{clccccccc}
        \toprule
        \makebox[0.03\textwidth][c]{} &
            \makebox[0.08\textwidth][l]{Method} & \makebox[0.1\textwidth][c]{Visual Backbone} & \makebox[0.1\textwidth][c]{Fold-0} & \makebox[0.09\textwidth][c]{Fold-1} & \makebox[0.09\textwidth][c]{Fold-2} & \makebox[0.09\textwidth][c]{Fold-3} & \makebox[0.09\textwidth][c]{Mean} & \makebox[0.09\textwidth][c]{FB-IOU}\tabularnewline
        \midrule
    \multirow{14}[4]{*}{\begin{sideways}1-shot\end{sideways}}     & PANet~\cite{PANet} & \multirow{3}{*}{ResNet-50} & 44.0  & 57.5  & 50.8  & 44.0    & 49.1  & - \tabularnewline

      & PGNet~\cite{PGNet} &            & 56.0    & 66.9  & 50.6  & 50.4  & 56.0    & 69.9 \tabularnewline
     & BAM~\cite{BAM}   &       & 69.0 & 73.6 & \textcolor[rgb]{ 0,  0,  1}{\textbf{67.6}} & 61.1 & 67.8 & - \tabularnewline
        \cmidrule{2-9}
       & FWB~\cite{FWB}   & \multirow{9}{*}{ResNet-101}  & 51.3  & 64.5  & 56.7  & 52.2  & 56.2  & - \tabularnewline
       & PPNet~\cite{PPNet} &          & 52.7  & 62.8  & 57.4  & 47.7  & 55.2  & 70.9 \tabularnewline
       & DAN~\cite{DAN}   &          & 54.7  & 68.6  & 57.8  & 51.6  & 58.2  & 71.9 \tabularnewline
       & PFENet~\cite{PFENet} &           & 60.5  & 69.4  & 54.4  & 55.9  & 60.1  & 72.9 \tabularnewline
       & RePRI~\cite{REPREI} &            & 59.6  & 68.6  & 62.2  & 47.2  & 59.4  & - \tabularnewline
       & HSNet~\cite{HSNet} &           & 67.3  & 72.3  & 62.0    & 63.1  & 66.2  & 77.6 \tabularnewline
       & IPMT~\cite{IPMT}  &            & 71.6  & 73.5  & 58.0    & 61.2  & 66.1  & - \tabularnewline
       & IPRNet~\cite{IPRNet} &            & 67.8  & \textcolor[rgb]{ 0,  0,  1}{\textbf{74.6}} & 65.7  & 62.2  & 67.5  & - \tabularnewline
      &  DACM~\cite{DACM}  &             & 68.7  & 73.5  & 63.4  & 64.2  & 67.5  & \textcolor[rgb]{ 0,  0,  1}{\textbf{78.9}} \tabularnewline
        \cmidrule{2-9}
       & DCAMA~\cite{DCAMA} & Swin-B  & 72.2  & 73.8  & 64.3  & \textcolor[rgb]{ 0,  0,  1}{\textbf{67.1}} & \textcolor[rgb]{ 0,  0,  1}{\textbf{69.3}} & 78.5 \tabularnewline
    &FPTrans~\cite{FPtrans} & DeiT-B/16      & \textcolor[rgb]{ 0,  0,  1}{\textbf{72.3}} & 70.6  & \textcolor[rgb]{ 1,  0,  0}{\textbf{68.3}} & 64.1  & 68.8  & - \tabularnewline
    \midrule
       & \textbf{ClsCLIP+ (Ours)} &   ViT-B/16      & \textcolor[rgb]{ 1,  0,  0}{\textbf{75.4}} & \textcolor[rgb]{ 1,  0,  0}{\textbf{79.2}} & 60.4 & \textcolor[rgb]{ 1,  0,  0}{\textbf{71.1}} & \textcolor[rgb]{ 1,  0,  0}{\textbf{71.5}} & \textcolor[rgb]{ 1,  0,  0}{\textbf{80.6}} \tabularnewline
    \bottomrule
     \end{tabular}
    }

    \vspace{-0.5cm}
    \label{table4}
\end{table*}

\section{Experiments}
\label{section:Experiments}

Follow CLIP-based one-stage ZS3 method~\cite{Lseg}, we compare our propose method with zero-shot and few-shot semantic segmentation models on the few-shot benchmark which is consistent with ``inductive'' settings.
Although few-shot semantic segmentation contains more known information about unseen classes than zero-shot, our method even outperforms few-shot semantic segmentation.

\subsection{Experiments setup}
\textbf{Dataset.}
We evaluate our method using the few-shot evaluation protocol on two widely used benchmarks: PASCAL-$5^i$~\cite{PASCAL5i} and COCO-$20^i$~\cite{COCO20i}.
PASCAL-$5^i$~\cite{PASCAL5i} is derived from PASCAL VOC 2012~\cite{VOC2012} with additional annotations from SDS~\cite{SDS}. 
It contains 20 classes divided into four folds: \{PASCAL-$5^i$: i = 0,1,2,3\}. 
COCO-$20^i$~\cite{COCO20i} is derived from COCO~\cite{COCO}, which contains 80 classes divided into four folds: \{COCO-$20^i$: i = 0,1,2,3\}.

\textbf{Implementation details.}
We use CLIP-ViT-B/16 as the backbone for both visual and text encoders in our proposed method. 
We use a lightweight decoder built with two-layer transformers.
For the optimizer, we adopt AdamW~\cite{ADamW} with a cyclic cosine annealing learning rate schedule~\cite{Loshchilov2016SGDRSG} and an initial learning rate of 0.001. 
We train our model on a single NVIDIA Tesla A100 for 200 epochs, using a batch size of 64 for PASCAL-$5^i$ and 512 for COCO-$20^i$. 


\textbf{Evaluation metrics.} 
We evaluate our method using two metrics: mean intersection over union (mIoU) and foreground-background intersection of union (FB-IoU), which are commonly used for evaluation in few-shot segmentation. 
mIoU computes the average IoU of all classes in a fold, i.e. $mIoU=\frac{1}{C} \sum_{c=1}^C IoU_c$, where C is the number of categories in a fold and $IoU_c$ is the IoU of each class. FB-IoU calculates the average IoU of foreground and background, ignoring the object class, i.e. FB-IoU = $\frac{1}{2}(IoU_F + IoU_B)$, where $IoU_F$ and $IoU_B$ represent the foreground and background IoU, respectively.
Normally, mIoU is more interesting than FB-IoU because it reflects how well the model generalizes to unseen classes. \par

\begin{figure}[t]
	\centering
	\includegraphics[width=\linewidth]{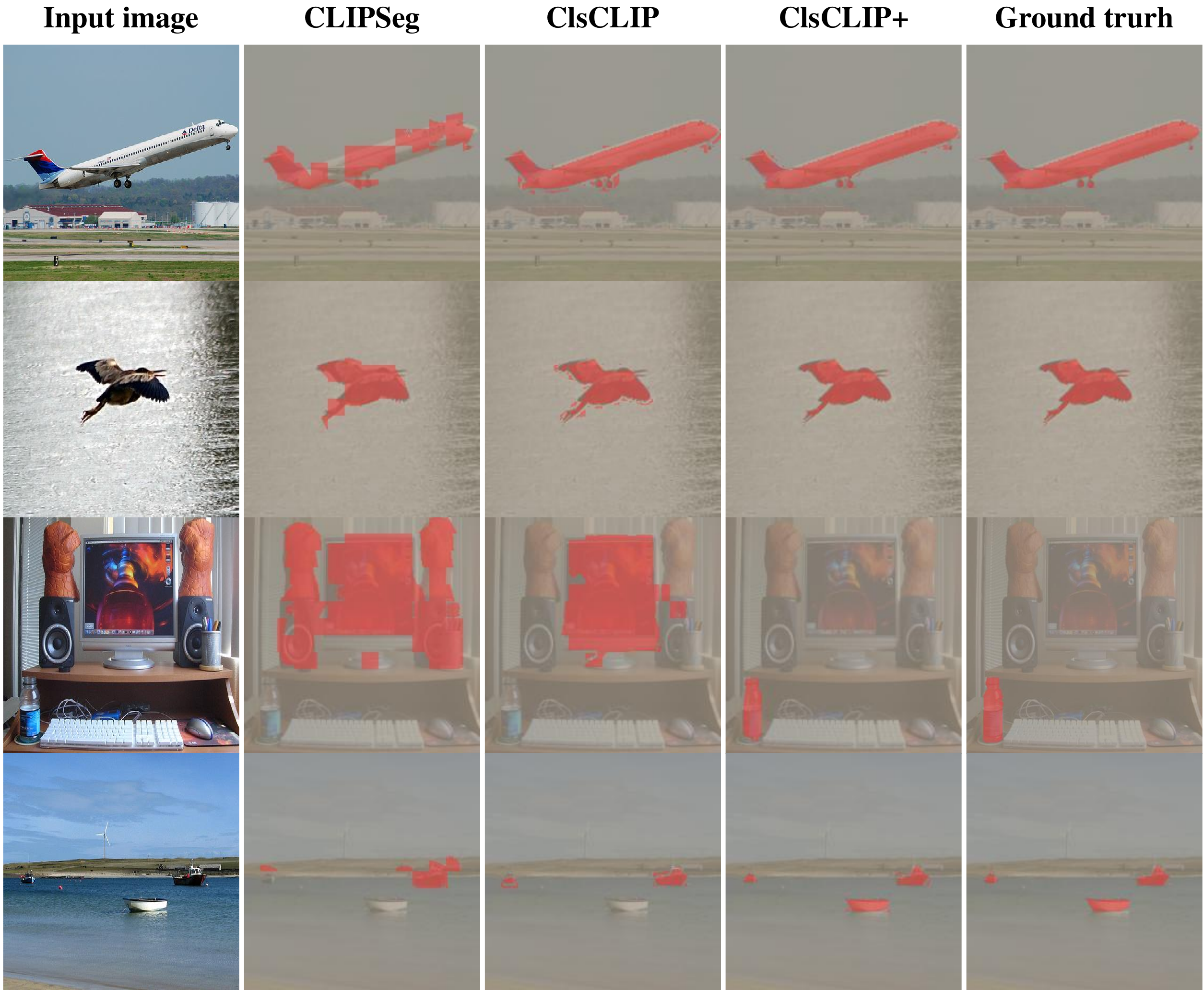}
	\caption{Comparison of prediction results of different methods. For larger objects, such as aeroplanes in the first row and birds in the second row, ClsCLIP performs better than CLIPSeg, while ClsCLIP+ performs more delicately. For tiny objects, such as bottles in the third row and boats in the fourth row, ClsCLIP and CLIPSeg are prone to missegmentation and missed segmentation, while ClsCLIP+ shows a strong effect on tiny object segmentation.}
	\label{fig5}
  \vspace{-0.5cm}
\end{figure}

\subsection{Results and analysis}

We first compare our ClsCLIP to the other ZS3 methods following the standard protocol, and report the available results shown in their papers. For the one-stage ZS3 method of CLIPSeg~\cite{clipseg}, we use the official code to obtain results. The typical CLIP-based two-stage approach~\cite{zsseg} will be evaluated and compared in detail in the supplementary material. In addition, we also compare our ClsCLIP+ to the few-shot methods following the standard protocol. Note that, our ClsCLIP+ still uses the zero-shot setting for comparison. 

Table \ref{table4} and Table \ref{table5} show the evaluation results on PASCAL-$5^i$ and COCO-$20^i$, respectively. 
ClsCLIP achieves the best performance compared to zero-shot methods, achieving 4.7 \% and 2.0 \% gains in mIoU scores than the suboptimal method of Lseg~\cite{Lseg} and CLIPSeg~\cite{clipseg} on the two datasets, respectively. In particular, although our proposal ClsCLIP+ is a zero-shot method, it achieves better results than the previous optimal 1-shot method of DCAMA~\cite{DCAMA} with more supervision, achieving 2.2 \% and 2.0 \% gains in mIoU scores. \par

\begin{table*}[t]
    \centering
    \small
    \caption{Evaluation results of mIoU and FB-IoU on COCO-$20^i$, \textcolor[rgb]{ 1,  0,  0}{\textbf{red}} and \textcolor[rgb]{ 0,  0,  1}{\textbf{blue}} indicate the best two scores. \textbf{Note that both ClsCLIP and ClsCLIP+ are zero-shot semantic segmantation methods.} ClsCLIP+ \textbf{with zero-shot setting} is even better than other 1-shot segmentation methods.}
     {(a). \textbf{Zero-shot method} ClsCLIP vs. Other zero-shot methods.}{
    \begin{tabular}{clccccccc}
        \toprule
        \makebox[0.03\textwidth][c]{} &
            \makebox[0.08\textwidth][l]{Method} & \makebox[0.1\textwidth][c]{Visual Backbone} & \makebox[0.1\textwidth][c]{Fold-0} & \makebox[0.09\textwidth][c]{Fold-1} & \makebox[0.09\textwidth][c]{Fold-2} & \makebox[0.09\textwidth][c]{Fold-3} & \makebox[0.09\textwidth][c]{Mean} & \makebox[0.09\textwidth][c]{FB-IOU}\tabularnewline
        \midrule
    
  \multirow{4}{*}{\begin{sideways}zero-shot\end{sideways}} &  ZS3Net& \multirow{1}{*}{ResNet101} & 18.8  & 20.1  & 24.8  & 20.5  & 21.1  & 55.1\tabularnewline

 &  LSeg~\cite{Lseg}  & ViT-L/16       & 28.1  & 27.5  & 30.0    & 23.2  & 27.2  & 59.9\tabularnewline
&   CLIPSeg~\cite{clipseg} & ViT-B/16       & \textcolor[rgb]{ 1,  0,  0}{\textbf{34.2}} & \textcolor[rgb]{ 0,  0,  1}{\textbf{38.9}} & \textcolor[rgb]{ 0,  0,  1}{\textbf{34.9}}  & \textcolor[rgb]{ 0,  0,  1}{\textbf{31.9}} & \textcolor[rgb]{ 0,  0,  1}{\textbf{35.0}} & \textcolor[rgb]{ 0,  0,  1}{\textbf{62.5}} \tabularnewline
\cmidrule{2-9}
  &  \textbf{ClsCLIP (Ours)} & ViT-B/16  & \textcolor[rgb]{ 0,  0,  1}{\textbf{34.1}} & \textcolor[rgb]{ 1,  0,  0}{\textbf{40.0}} & \textcolor[rgb]{ 1,  0,  0}{\textbf{38.2}} & \textcolor[rgb]{ 1,  0,  0}{\textbf{35.7}} & \textcolor[rgb]{ 1,  0,  0}{\textbf{37.0}} & \textcolor[rgb]{ 1,  0,  0}{\textbf{64.5}}\tabularnewline
\bottomrule
 
    \end{tabular}
    }
      
     \vspace{0.2cm}
    {(b). \textbf{Zero-shot method} ClsCLIP+ vs. Other 1-shot methods.}{
   
    \begin{tabular}{clccccccc}
        \toprule
        \makebox[0.03\textwidth][c]{} &
            \makebox[0.08\textwidth][l]{Method} & \makebox[0.1\textwidth][c]{Visual Backbone} & \makebox[0.1\textwidth][c]{Fold-0} & \makebox[0.09\textwidth][c]{Fold-1} & \makebox[0.09\textwidth][c]{Fold-2} & \makebox[0.09\textwidth][c]{Fold-3} & \makebox[0.09\textwidth][c]{Mean} & \makebox[0.09\textwidth][c]{FB-IOU}\tabularnewline
        \midrule
    \multirow{14}[4]{*}{\begin{sideways}1-shot\end{sideways}}    &PPNet~\cite{PPNet} & \multirow{5}[2]{*}{ResNet50}  & 28.1  & 30.8  & 29.5  & 27.7  & 29.0    & - \tabularnewline

    &   RPMM~\cite{RPMM}  &             & 29.5  & 36.8  & 28.9  & 27.0    & 30.6  & - \tabularnewline
    & RePRI~\cite{REPREI} &          & 32.0    & 38.7  & 32.7  & 33.1  & 34.1  & - \tabularnewline
    & BAM~\cite{BAM}   &           & 43.4 & 50.6 & 47.5 & 43.4 & 46.2 & - \tabularnewline
   & DACM~\cite{DACM}  &         & 37.5  & 44.3  & 40.6  & 40.1  & 40.6  & 68.9 \tabularnewline
        \cmidrule{2-9}
       & FWB~\cite{FWB}   & \multirow{6}[2]{*}{ResNet101}  & 17.0    & 18.0    & 21.0    & 28.9  & 21.2  & -  \tabularnewline
       & DAN~\cite{DAN}   &         & -     & -     & -     & -     & 24.4  & 62.3 \tabularnewline
        & PFENet~\cite{PFENet} &            & 36.8  & 41.8  & 38.7  & 36.7  & 38.5  & 63.0  \tabularnewline
      &     HSNet~\cite{HSNet} &          & 37.2  & 44.1  & 42.4  & 41.3  & 41.2  & 69.1\tabularnewline
      &    IPMT~\cite{IPMT}  &            & 40.5  & 45.7  & 44.8  & 39.3  & 42.6  & -\tabularnewline
   &     IPRNet~\cite{IPRNet} &            & 42.9  & 50.6  & 46.8  & 47.4  & 46.9  & - \tabularnewline
        \cmidrule{2-9}
     &  DCAMA~\cite{DCAMA} & Swin-B  & \textcolor[rgb]{ 0,  0,  1}{\textbf{49.5}}  & \textcolor[rgb]{ 1,  0,  0}{\textbf{52.7}} & \textcolor[rgb]{ 1,  0,  0}{\textbf{52.8}} & \textcolor[rgb]{ 0,  0,  1}{\textbf{48.7}}  & \textcolor[rgb]{ 1,  0,  0}{\textbf{50.9}} & -  \tabularnewline
  &  FPTrans~\cite{FPtrans} & DeiT-B/16       & 44.4  & 48.9  & 50.6  & 44.0    & 47.0    & - \tabularnewline
    \midrule
       & \textbf{ClsCLIP+ (Ours)} & ViT-B/16   & \textcolor[rgb]{ 1,  0,  0}{\textbf{55.0}} & \textcolor[rgb]{ 0,  0,  1}{\textbf{49.9}} & \textcolor[rgb]{ 0,  0,  1}{\textbf{54.6}} & \textcolor[rgb]{ 1,  0,  0}{\textbf{52.2}} &\textcolor[rgb]{1,  0,  0}{\textbf{52.9}} & \textcolor[rgb]{ 1,  0,  0}{\textbf{73.9}}  \tabularnewline  
       \bottomrule
       \end{tabular}
    }
    \label{table5}
      \vspace{-0.4cm}
\end{table*}
To demonstrate the effectiveness of our proposed method for segmentation tasks, we compare the prediction results of our method and CLIPSeg on various images. 
Figure~\ref{fig5} illustrates that ClsCLIP performs better than CLIPSeg, but both methods struggle to segment tiny objects accurately. 
By incorporating a region proposal pre-processing, we enhance ClsCLIP to handle tiny object segmentation better.
\begin{table}[t]
  \centering
   \small
   
  \caption{Evaluation results using YOLO directly on PASCAL-$5^i$.}
  \vspace{2mm}
    \begin{tabular}{ccc}
    \toprule
     \makebox[0.13\textwidth][c]{Method} & \makebox[0.13\textwidth][c]{mIoU (\%)} & \makebox[0.14\textwidth][c]{FB-IOU (\%)}\tabularnewline
 \midrule
    YOLO  & 55.5  & 71.8  \tabularnewline
    ClsCLIP+ & \textbf{71.5} & \textbf{80.6}  \tabularnewline
\bottomrule
    \end{tabular}%
    \label{table6}
      \vspace{-0.2cm}
\end{table}%
To address a potential concern that the high mIoU score of ClsCLIP+ is mainly due to the quality of the regions generated by YOLO, we conduct additional experiments to validate our method. 
Table \ref{table6} shows that using the YOLO regions as the segmentation results can achieve some improvement, but it is still far behind ClsCLIP+ (55.5\% vs. 71.5\% in mIoU score), which demonstrates the effectiveness of our one-way [CLS] token navigation.\par

To explore the upper bound of our proposed method, we evaluate ClsCLIP’s predictions under the assumption that the region proposals generated in the pre-processing are accurate and complete. 
We use a subset of fold0 from PASCAL-$5^i$ (sub-fold0), which has test images with only 26.2\% mIoU using YOLO as the region proposal generator.
We replace YOLO with manual labels to ensure that all region proposals are correct. 
Table~\ref{table7} shows that ClsCLIP achieves a significant improvement in mIoU score on sub-fold0, which obtains 23.6\%  gains in mIoU score by using manually annotated proposals.
This suggests that ClsCLIP’s prediction accuracy will increase with the precision of the prior localization information, further proving the effectiveness of our one-way [CLS] token navigation.

\begin{table}[t]
  \centering
   \small
  \caption{Evaluation results of YOLO and Manual annotation.}
\vspace{2mm}
    \begin{tabular}{cc}
    \toprule
    \makebox[0.21\textwidth][c]{Method} & \makebox[0.21\textwidth][c]{mIoU (\%)}\tabularnewline
    \midrule
    YOLO  & 26.2  \tabularnewline
    Manual annotation & \textbf{49.8 \textcolor{my_green}{(+23.6)}} \tabularnewline
    \bottomrule
    \end{tabular}%
  \label{table7}%
  \vspace{-0.4cm}
\end{table}%
\section{Conclusion}

In this work, we propose a one-way [CLS] token navigation to extend the capabilities of CLIP zero-shot classification to zero-shot semantic segmentation, dubbed ClsCLIP.
Concretely, ClsCLIP replaces the [CLS] token in shallow layers of the visual encoder with the text-side [CLS] tokens to embed global category prior information and pay more attention to the region of interest. 
Furthermore, we employ a region proposal pre-processing to strengthen ClsCLIP to alleviate the issue that the segmentor is prone to missing tiny objects.  
Extensive experiments demonstrate that our proposed  method achieves a state-of-the-art performance with other zero-shot semantic segmentation methods, and it is even comparable with those few-shot methods.

{\small
\bibliographystyle{ieee_fullname}
\bibliography{egbib}
}

\end{document}